\pdfoutput=1

\documentclass[11pt]{article}

\usepackage[]{acl}

\usepackage{times}
\usepackage{latexsym}

\usepackage[T1]{fontenc}

\usepackage[utf8]{inputenc}

\usepackage{microtype}

\usepackage{inconsolata}
\usepackage[ruled,vlined,linesnumbered]{algorithm2e}
\SetAlFnt{\small}
\SetAlCapFnt{\small}
\SetAlCapNameFnt{\small}
\usepackage{algorithmic}
\usepackage{float}
\usepackage{multirow}
\usepackage{booktabs}

\usepackage{amsmath}
\usepackage{amssymb}

\usepackage{kky}
\usepackage{cleveref}
\newcommand{\Scref}[1]{\S\ref{#1}}
\usepackage{tcolorbox}
\usepackage{tabularx}

\usepackage{booktabs} 
\usepackage{multicol}
\usepackage{enumitem}
\usepackage{caption}
\usepackage{subcaption}
\usepackage{xspace}
\usepackage{bbm}
\newcommand{\onlinebuf}{BST\xspace}
\newcommand{\onlinecum}{CST\xspace}
\newcommand{\offline}{AST\xspace}
\newcommand{\tunesource}{Src-Only\xspace}

\newcommand{\tunetrue}{Supervised\xspace}

\newcommand{\promptzero}{Zero\xspace}

\usepackage[textsize=scriptsize]{todonotes}

\definecolor{darkyellow}{HTML}{FDBC42}

\title{Evolving Domain Adaptation of Pretrained Language Models for Text Classification}

\author{
Yun-Shiuan Chuang$^1$ \ \ \ Yi Wu$^1$ \ \ \ Dhruv Gupta$^1$ \ \ \ Rheeya Uppaal$^1$ \\
\textbf{Ananya Kumar$^2$ \ \ \ Luhang Sun$^1$ \ \ \ Makesh Narsimhan Sreedhar$^1$ \ \ \ Sijia Yang$^1$} \\
\textbf{Timothy T. Rogers$^1$ \ \ \ Junjie Hu$^1$ }
\\
$^1$University of Wisconsin-Madison \ \ \ $^2$Stanford University \ \ \  
\\
\small \texttt{\{yunshiuan.chuang, ywu676, dgupta42, uppaal, lsun232, msreedhar, syang84, ttrogers,junjie.hu\}@wisc.edu} \ \ \ \\
\small \texttt{ananya@cs.stanford.edu} 
}

\begin{document}
\maketitle

\begin{abstract}
Adapting pre-trained language models (PLMs) for time-series text classification amidst evolving domain shifts (EDS) is critical for maintaining accuracy in applications like stance detection. This study benchmarks the effectiveness of evolving domain adaptation (EDA) strategies, notably self-training, domain-adversarial training, and domain-adaptive pretraining, with a focus on an incremental self-training method. Our analysis across various datasets reveals that this incremental method excels at adapting PLMs to EDS, outperforming traditional domain adaptation techniques. These findings highlight the importance of continually updating PLMs to ensure their effectiveness in real-world applications, paving the way for future research into PLM robustness against the natural temporal evolution of language.
\end{abstract}


\section{Introduction}
\label{sec:introduction}

Text classification using pre-trained language models (PLMs)~\cite{delvin2019bert,brown2020language} has been widely used in online applications over time, such as predicting public sentiments on Twitter or product review rating on Amazon. A common challenge in these applications is the continuous evolution of data over time, leading to significant changes in data distribution~\cite{aldayel2021stance,kuccuk2020stance}. This evolving nature results in notable performance degradation of PLMs, as they struggle to adapt to new, unseen data distributions. Despite this challenge, the rapid annotation of massive time-series data to mitigate the model degradation is often infeasible, pushing the necessity to adapt PLMs effectively using up-to-date unlabeled data.


Building on the challenges highlighted previously, this leads us to the formal introduction of the concept of \textit{evolving domain shift} (EDS)~\cite{alkhalifa2021opinions,alkhalifa2022capturing,mu2023examining}. EDS characterizes the gradual divergence in data distribution, a phenomenon particularly pronounced in time-series text classification. For instance, in the context of COVID-19 vaccination, a stance classifier (i.e., determining if a tweet is in favor of or against vaccination) can be trained to assess public opinions on vaccines. A BERT model~\cite{delvin2019bert}, trained with tweets from the first six months of the vaccine debate (December 2020 to May 2021), initially achieves a macro F1 score $F_{\text{macro}}=0.654$. However, its performance degrades steadily over time. In the subsequent 13 months (June 2021 to June 2022), the macro F1 score averaged across months drops to $F_{\text{avg}}=0.509$, underscoring a significant decline in classifier performance. This dynamic nature of data necessitates continuous adaptation to new information and shifting public sentiment as new virus variants and vaccine brands emerge. To effectively address EDS, we explore a series of \textit{evolving domain adaptation} (EDA) techniques to adapt PLMs in accordance with these ongoing shifts, aiming to mitigate such performance degradation over time.

\begin{table*}[tbp!]
  \centering  
  \small
  \resizebox{\linewidth}{!}{  
    \begin{tabular}{llllllll}
      \toprule
      & & \multicolumn{4}{c}{BERT-large} & \multicolumn{2}{c}{FLAN-T5-XXL} \\ 
      \cmidrule(lr){3-6} \cmidrule(lr){7-8}
      Methods & Variants & COVID & WTWT & SCIERC & PUBCLS & COVID & SCIERC \\
      \midrule
      Baseline & \tunesource & 0.509 & 0.618 & 0.485 & 0.406 & 0.795  &  0.809 \\
      \midrule       
      DANN & - & 0.519 (6$\uparrow$) & 0.647 (17$\uparrow$) & 0.511 (8$\uparrow$) & 0.440 (14$\uparrow$) & - & - \\
      DAPT & - & 0.524 (8$\uparrow$) & 0.664 (28$\uparrow$) & 0.503 (5$\uparrow$) & 0.458 (21$\uparrow$) & 0.731 (<0)  & 0.503 (<0) \\      
      \midrule       
      Self-training & \offline & 0.460 (28$\downarrow$) & 0.633 (9$\uparrow$) & 0.593 (32$\uparrow$) & 0.363 (18$\downarrow$) & 0.816 (46$\uparrow$)  &  0.820 (22$\uparrow$) \\    
       & \onlinebuf  & 0.569 (34 $\uparrow$) & 0.697 (47$\uparrow$) & 0.520 (10$\uparrow$) & 0.469 (26$\uparrow$) & 0.812 (37$\uparrow$) &  0.822 (39$\uparrow$) \\
       & \onlinecum  & \textbf{0.611 (57$\uparrow$)} & \textbf{0.739 (72$\uparrow$)} & \textbf{0.684 (59$\uparrow$)} & \textbf{0.492 (35$\uparrow$)} & \textbf{0.821 (57$\uparrow$)}  & \textbf{0.842 (100$\uparrow$)} \\
      \midrule    
      \tunetrue & - & 0.687 & 0.785 & 0.822 & 0.651 & 0.841 & 0.842 \\    
      \midrule
      \promptzero & - & - & - & - & - & 0.772  &  0.348 \\
      \bottomrule
    \end{tabular}
  }
  \caption{Performance metrics across methods for the BERT-large and FLAN-T5-XXL models on all datasets. Each entry shows $F_\text{avg}$, as well as $\Delta_{\text{avg,norm}}$ for each adaptive method, shown in percentage in the parentheses (\%).}
  \label{tab:performance_metrics_tuning_merged}
  \vspace{-3mm}
\end{table*}

In this work, we focus on exploring and comparing variants of domain adaptation techniques using pre-trained language models in the context of evolving domain shifts. Specifically, we evaluate three common domain adaptation approaches: \textit{self-training (ST)}~\cite{amini2022self}, \textit{domain-adversarial training (DANN)}~\cite{ganin2016domain}, and \textit{domain-adaptive pretraining (DAPT)}~\cite{luu2022time,gururangan2020don}. For self-training, inspired by the concept of ``gradual self-training''~\cite{kumar2020understanding}, we propose an incremental self-training method by caching pseudo-labeled data in a dynamic buffer over time. Specifically, this method iteratively predicts labels for unlabeled data and fine-tunes the language model on the evolving pseudo-labeled dataset in the buffer. Crucially, we incorporate upsampling in our self-training approach to address label shift that naturally occurs in evolving domains. This step is vital to rebalance the class distribution in our datasets, ensuring that our models remain effective and accurate over time. Additionally, we compare popular PLMs in different model architectures with different scales, including an encoder-only \textit{discriminative model} and a state-of-the-art encoder-decoder \textit{generative} large language model (LLM).

We conduct experiments on four real-world text classification datasets in public health, finance, scientific publications, and news domains. These datasets exhibit significant evolving domain shifts. Our findings indicate that incremental self-training surpasses unadapted baselines and other domain adaptation methods across all four datasets. For example, for the COVID-19 vaccination dataset, while unadapted baseline has $F_{\text{avg}}=0.509$ on the 13 months with unlabeled tweets, DANN has $F_{\text{avg}}=0.519$ and DAPT has $F_{\text{macro}}=0.524$, self-training in an incremental way reaches $F_{\text{avg}}=0.611$, outperforming all methods. Similar trends hold for the other datasets as well. Notably, we conducted an ablation study to verify the critical role of upsampling in our approach. This study confirms that upsampling is important in rebalancing class distribution to mitigate label shift.



In summary, our contributions are three-fold:
\begin{itemize}[leftmargin=13pt]\itemsep-0.2em
    \item We systematically compare the evolving domain adaptation ability of popular PLMs in various scales and architectures for text classification across four domains, and evaluation different self-training strategies of PLMs. 
    \item We propose an incremental self-training method with a dynamic buffer and demonstrate its effectiveness over other adaptation methods like DAPT and DANN.
    \item We collect a stance detection dataset that reflects the evolving domain shift of tweets on COVID-19 vaccination over a time span of 1.5 years since the start of COVID-19 vaccination debate, enabling future developments on the realistic EDA task.
\end{itemize}
\section{Preliminaries}
\label{sec:prelim}

\paragraph{Evolving Domain Adaptation (EDA):} In EDA, we address a $M$-way text classification task over time. We have an initial labeled dataset $\Dcal_0$ from distribution $P_{\mathcal{XY}}^{0}$ and a series of subsequent unlabeled datasets $(\Dcal_1, \Dcal_2, \dots, \Dcal_T)$ each from a unique input distribution $P_{\mathcal{X}}^{t}$. There's an evolving shift in the input distribution across consecutive time steps, signified by $P_{\mathcal{X}}^{t-1} \neq P_{\mathcal{X}}^{t}$. Label distribution, $P_{\Ycal}^t$, also evolves over time, implying $P_{\Ycal}^{t-1}\neq P_{\Ycal}^{t}$. The dataset concatenation between two time steps is denoted as $\mathcal{D}_{t:t'}$. EDA's objective is leveraging all datasets $\mathcal{D}_{0:T}$ to optimize a classifier $f_\theta: \mathcal{X} \to \mathcal{Y}$ across all unlabeled domains from $t\in [1,T]$. 

\begin{figure}[tbp!]
\centering
\includegraphics[width=1\linewidth]{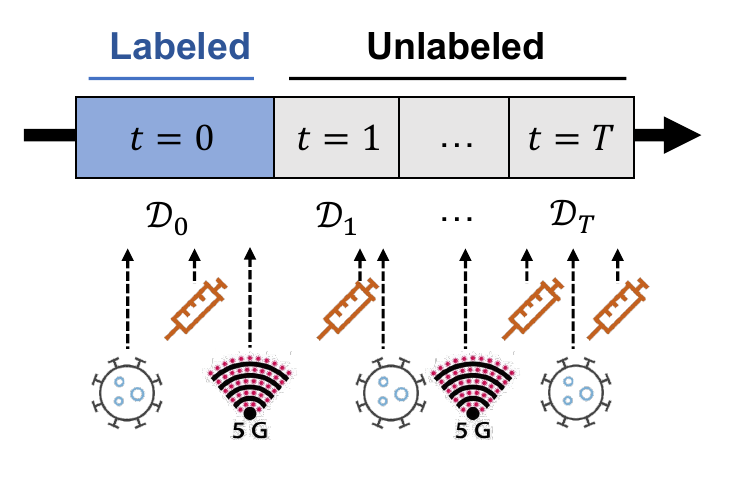}
\caption{In evolving domain adaptation, we start with labeled data from a source domain and progressively incorporate unlabeled data from a series of target domains. These target domains evolve away from the source domain over time due to factors like the emergence of new virus variants, vaccine brands, or conspiracy theories.}
\label{fig:gda_schematic}
\end{figure}

\section{Methods}
\label{sec:method}

We consider three popular domain adaptation methods, including self-training methods (ST), domain-adaptive pretraining (DAPT), and domain-adversarial neural networks (DANN). We also analyze their time and space complexity regarding the number of target domains, denoted as $T$.

\subsection{Fine-tuning with Self-training}
\label{sec:method-self-training}
Self-training~\cite{self-training,amini2022self} (ST) is a common approach to semi-supervised learning, which leverages a base model to predict pseudo-labels for unlabeled data and further fine-tunes the model on pseudo-labeled data.

Denote a pre-trained language model encoder as $g: \mathcal{X}\to \mathbb{R}^d$ that embeds a text input $x$ into an embedding $\xb = g(x)$. We employ a randomly initialized prediction layer $h:\mathbb{R}^d\to \mathcal{Y}$ on top of the embeddings to predict the class label. Therefore, we have $f_\theta=g\circ h$, and we update the model parameters in both $g$ and $h$ during fine-tuning. This fine-tuning strategy has proven effective on encoder-only models (e.g., BERT) by a recent study~\cite{delvin2019bert}, we further extend this strategy to instruction-tuned encoder-decoder models (e.g. FLAN-T5-XXL) to generate label text strings without adding the prediction layer.

\paragraph{Buffered Self-Training (\onlinebuf)} In the \onlinebuf method, we maintain a dynamic buffer $\Bcal_t$ of fixed size $b$ at any time step $t\in [0,T]$. When $t=0$, the buffer is initialized with the source labeled dataset $\Bcal_0=\Dcal_{0}$, and a fixed size $b=|\Dcal_{0}|$. For the subsequent time steps $t\in [1, T]$, a new model $f_{\theta, t}$ is fine-tuned on the buffer $\Bcal_{t-1}$, and then used to generate the pseudo-labels for the unlabeled instances from $\Dcal_{t}$. Specifically, we obtain a pseudo-label $\tilde{y}_i=f_{\theta, t}(x_i)$ for each unlabeled example $x_i\in \Dcal_t$. The pseudo-labeled examples $\tilde{\Dcal}_t=\{(x_i,\tilde{y}_i)|x_i \in \Dcal_t\}$ are then added to the buffer $\Bcal_t$. To maintain a fixed-size buffer, we update the buffer by a First-In-First-Out (FIFO) strategy i.e. the addition of a new instance leads to the removal of the oldest one when the buffer is full. An instance in the buffer can either be a labeled instance from $\Dcal_0$ or a pseudo-labeled instance from $\tilde{\Dcal}_t$. This is represented as:

\begin{align}
\label{eq:buffer_fifo}
\Bcal_{t} = \text{FIFO}(\Bcal_{t-1} \cup \{(x_i, \tilde{y}_i) | x_i\in \Dcal_t\}, b)
\end{align}
where $\text{FIFO}(\cdot, b)$ is a function that maintains the most recent $b$ instances in the buffer. Notably, when fine-tuning $f_{\theta,t}$, the training set in $\Bcal_{t-1}$ is first upsampled to mitigate label imbalance. Therefore, regardless of label shift, there are always equal number of instances per class to fine-tune the model.

\begin{algorithm}
\DontPrintSemicolon
\KwIn{Source domain $\Dcal_0$, target domains $\Dcal_{1:T}$}
Initialize buffer $\Bcal_0 = \Dcal_0$, buffer size $b=|\Dcal_0|$\;
\For{$t = 1$ to $T$}{
    Train model $f_{\theta, t}$ on $\Bcal_{t-1}$ \;
    
    Obtain pseudo-label $\tilde{y}_i = f_{\theta, t}(x_i), \forall x_i \in D_{t}$ \;

    Update $\Bcal_{t}$ by FIFO in Eq.~(\ref{eq:buffer_fifo});
}
\KwOut{$\{f_{\theta, t}\}_{t=1}^{T}$}
\caption{Online Buffered Self-Training (\onlinebuf)}
\label{algo:stb}
\end{algorithm}


\paragraph{Cumulative Self-Training (\onlinecum)} The \onlinecum method differs from \onlinebuf in the management of the buffer. Instead of maintaining a fixed size buffer, \onlinecum uses a cumulative buffer that grows unbounded over time, also denoted as $\Bcal_t$ at the $t$-th step. Specifically, we replace the buffer update in line 5 of Algorithm \ref{algo:stb} as follows for \onlinecum:
\begin{align} \label{eq:onlinecum}
\Bcal_{t} = \Bcal_{t-1} \cup \{(x_i, \tilde{y}_i)|x_i\in \Dcal_t\}.
\end{align}
This update rule ensures that all pseudo-labeled instances are retained in the buffer across all time steps. The rest of the algorithm remains the same as in \onlinebuf, including training the model on the current buffer and the generation of pseudo-labels for data at the next time step. We propose \onlinecum to test the importance of accumulating historical data.

\paragraph{All Self-Training (\offline)} In the \offline method, we first train a base model $f_{\theta, 0}$ on the source domain $\Dcal_0$. We then use this model to generate pseudo-labels for all instances across the entire series of target domains $\Dcal_{1:T}$. Namely, we obtain a pseudo-label $\tilde{y}_i = f_{\theta, 0}(x_i)$ for each unlabeled instance $x_i \in \Dcal_{t}$ over all subsequent time steps $t \in [1,T]$. We denote the $t$-th pseudo-labeled dataset as $\tilde{\Dcal}_t=\{(x_i,\tilde{y}_i)|x_i \in \Dcal_t\}$. Next, we concatenate the initial source labeled dataset and the pseudo-labeled datasets from all target domains as $\tilde{\Dcal}_{0:T}' = \Dcal_0 \cup \tilde{\Dcal}_{1}, \dots, \tilde{\Dcal}_{T}$. Finally, we train a new model $f_{\theta, T}$ on this combined dataset $\tilde{\Dcal}_{0:T}'$. We propose \offline to test the if we can ignore the temporal information.

\paragraph{Complexity Analysis:}
\onlinebuf operates with linear time complexity $O(T)$ due to sequential processing and has a constant space complexity ($O(1)$), given the fixed buffer size. \onlinecum's time complexity is quadratic $O(T^2) $because of the growing training set with each domain, and its space complexity scales linearly $O(T)$ with the number of target domains. \offline has a linear time complexity of $O(T)$ for pseudo-labeling and training on all domains. Its space complexity is also $O(T)$ to store the pseudo-labeled data for all target domains prior to the second pass of training.

\subsection{Domain-Adaptive Pretraining (DAPT)}
\label{sec:method-continued-pretraining}
Domain-Adaptive Pretraining (DAPT) is a technique where a pretrained language model (PLM) is further pretrained on a domain-specific corpus to enhance its performance in that domain \cite{gururangan2020don}. DAPT has shown its effectiveness at mitigating temporal misalignment \cite{luu2022time}. In our EDA setting, we apply DAPT by continuing pretraining on each target domain $\Dcal_t$ before fine-tuning on the source domain $\Dcal_0$. Specifically, at each time step $t>0$, we initialize the model weights from the pretrained weights $f_{\theta,0}$. We then continue pretraining $f_{\theta,0}$ on the target domain $\Dcal_t$ with the pretraining loss (e.g., masked language modeling). After DAPT, we fine-tune the adapted model $f_{\theta,t}$ on labeled source data $\Dcal_0$ for classification. Through continued pretraining on each target domain, we aim to align the model temporally and adapt it to each target domain before fine-tuning. 

\paragraph{Complexity Analysis:} DAPT has a linear time complexity of $O(T)$ due to the necessity of performing the pretraining process individually for each of the $T$ target domains. The space complexity is $\text{max}(O(|\Dcal_{0}|), O(|\Dcal_{t}|))=O(1)$, where $|\Dcal_{t}|$ is the size of each target domain for pretraining, and $|\Dcal_{t}|$ for fine-tuning with source domain.

\subsection{Domain-Adversarial Training (DANN)}
\label{sec:method-dann}

Domain-adversarial training of neural networks (DANN) is an approach that promotes \textit{domain-invariant feature} learning by adversarial alignment of domain distributions \cite{ganin2016domain}. We extend DANN to our EDA setting by constructing a domain discriminator $d_\phi$ that is trained to distinguish between source domain $\Dcal_0$ and a given target domain $\Dcal_t$. Specifically, at each time step $t$, the feature extractor $g_\theta$ is trained adversarially against the domain discriminator $d_\phi$ to extract domain-invariant features. This is achieved by a gradient reversal layer between $g_\theta$ and $d_\phi$. The optimization objective contains a classification loss $\Lcal_{cls}$ that encourages correct label predictions, along with a domain adversarial loss $\Lcal_{adv}$: $\min_{\theta} \max_{\phi} \ \Lcal_{cls}(f_{\theta}, \Dcal_0) - \lambda \Lcal_{adv}(g_\theta, d_\phi, \Dcal_0, \Dcal_t)$.

\paragraph{Complexity Analysis:} For DANN, time complexity is $O(T)$, as adversarial training is applied once per domain. The space complexity is $O(|\Dcal_{0}|)+O(|\Dcal_{t}|)=O(1)$, with training set size fixed to the combined size of the source and current target domain.
\section{Experimental Settings}
\label{sec:setting}

\subsection{Datasets}

\paragraph{COVID-19 Vaccination Dataset (COVID)} is constructed by the authors (See Appendix~\ref{sec:appendix-covid}). It contains 5002 tweets about COVID-19 vaccination, sampled daily from December 1, 2020, to June 30, 2022. Tweets are categorized as '\textit{Against}' (anti-vaccine views) or '\textit{Not-against}' (pro-vaccine, neutral, or ambiguous views). The source domain combines the first six months, and the rest form ten subsequent target domains (See Appendix~\ref{sec:data-preprocess}). 
\paragraph{Will-They-Won’t-They Dataset (WTWT)} features tweets about Mergers and Acquisitions (M\&A) \citep{conforti2020will}. They are classified into '\textit{Support}', '\textit{Refute}', '\textit{Comment}', and '\textit{Unrelated}'. We use the subset of 44,717 tweets from June 2015 to December 2018. The first year is the source domain, and the rest form 14 subsequent target domains (See Appendix~\ref{sec:data-preprocess}).
\paragraph{Computer Science Paper Abstract Dataset (SCIERC)} contains 8089 scientific entities found within the abstracts of computer science papers published between 1980 and 2016 \cite{luan2018multi}. These entities are grouped into six distinct categories: task, method, metric, material, other-scientific-term, and generic. Following the time partitions in \citet{luu2022time}, we designate the from 1980-1999 as the source domain, and the rest as a series of target domains.
\paragraph{Publisher Classification Dataset (PUBCLS)} comprises approximately 67k documents published by various publishers between 2009 and 2016 \cite{card2015media}. The objective is to classify each document based on its publisher. Following the time partitions in \citet{luu2022time}, we designate documents from 2009-2010 as the source domain, and the rest as a series of target domains.

\subsection{Performance Evaluation Metric}

Model performance is assessed on target domain test sets $\Dcal_{1:T}$. For each $t \in [1,T]$, the macro-averaged F1 Score, $F_{\text{macro},t}$, is computed over the label space $\Ycal$. The global performance metric, $F_\text{avg}$, is the average of all $F_{\text{macro},t}$, defined as $F_\text{avg} = \frac{1}{T}\sum_{t=1}^{T}F_{\text{macro},t}$.

In addition, we introduce the \textit{relative gain} metric to assess the relative improvement of each method over the baseline. This metric, denoted as $\Delta_{\text{avg,norm}}$, is computed as follows:$\Delta_{\text{avg,norm}} = (F_{\text{avg,method}} - F_{\text{avg},\tunesource}) / (F_{\text{avg},\tunetrue} - F_{\text{avg},\tunesource})$. Here, $F_{\text{avg,method}}$ is the $F_\text{avg}$ score of the method under evaluation, $F_{\text{avg},\tunesource}$ is the $F_\text{avg}$ score of the upadapted baseline \tunesource, and $F_{\text{avg},\tunetrue}$ is the $F_\text{avg}$ of the fully-supervised baseline \tunetrue (see \Scref{sec:baselines} for the definitions of the baselines). This metric $\Delta_{\text{avg,norm}}$ is intended to quantify the percentage improvement of a method relative to the baseline.

\subsection{Fine-tuning Language Models}
\label{sec:method-plms}

We explore two types of large PLMs - discriminative and generative models. For the discriminative model, we use \textbf{BERT-large-uncased} (BERT), a bidirectional transformer encoder \cite{delvin2019bert}. BERT embeddings are fed to a classification head fine-tuned with cross-entropy loss for text classification. 

For the generative model, we use \textbf{FLAN-T5-XXL}, a state-of-the-art, instruction-finetuned encoder-decoder LLM \cite{chung2022scaling, ziems2023can} \footnote{Among the model variants, we choose the FLAN-T5-XXL, as it demonstrates the strong zero-shot and few-shot performance in stance detection task \cite{ziems2023can}.}. We fine-tune FLAN-T5-XXL using casual language modeling loss, where both the input (the tweet embedded in a \textit{prompt $\Tcal$}) and the output is text (the label in verbal description). To make it computationally feasible, we fine-tune FLAN-T5-XXL using low-rank adaptation (LoRA)~\cite{hu2022lora}. Please refer to Table~\ref{tab:prompt-template-zero} for the prompt used for each dataset. Due to the compute constraint, we only evaluate FLAN-T5-XXL with the two smaller datasets COVID and SCIERC.

\subsection{Baselines}\label{sec:baselines}
 \textbf{Source-only Baseline (\tunesource):} The source-only baseline, $f_{\theta,0}$, is trained solely on the source domain $\Dcal_0$, without considering any target domain data. After training, this model is directly tested on all subsequent target domains, $\Dcal_{1:T}$. This baseline serves as a unadapted benchmark. \textbf{Fully-supervised Baseline (\tunetrue):} The fully-supervised baseline (\tunetrue) has  access to the true labels from all the target domains. This method trains on the fully labeled data from both the source and target domains. As such, it offers an upper bound on the performance achievable by EDA.

\subsection{Self-Training Methods}

\textbf{\onlinebuf, \onlinecum, \offline:} as detailed in \Scref{sec:method-self-training}. 
For training, validation, and testing partitions, please see Appendix~\ref{sec:appendix-train_vali_partition}. Hyperparameters for fine-tuning BERT-large and FLAN-T5-XXL are described in Appendix~\ref{sec:appendix-tuning_hyperparams}. 

\subsection{Domain-Adaptive Pretraining (DAPT)}As detailed in \Scref{sec:method-continued-pretraining}, DAPT can be divided into two stages: continued pretraining with the target domain data and then fine-tuning with source domain labeled data. For continued pretraining BERT, we use the mask language modeling (MLM) loss \cite{delvin2019bert}, whereas for continued pretraining FLAN-T5-XXL, we use causal language modeling (LM) loss with LoRA~\cite{hu2022lora}.

\subsection{Domain-Adversarial Training (DANN)}

 Hyperparameters are detailed in Appendix~\ref{sec:appendix-tuning_hyperparams}. Note that we only do DANN for BERT as it is unclear how DANN would work for a generative model like FLAN-T5-XXL.
\section{Results and Analysis}
\label{sec:results}

\paragraph{Incremental self-training outperforms DAPT and DANN}


Our comprehensive analysis showcases the superior performance of incremental self-training methods, both the buffered approach \onlinebuf and the cumulative approach \onlinecum, in the Evolving Domain Adaptation (EDA) setting. As shown in Table \ref{tab:performance_metrics_tuning_merged}, \onlinecum consistently surpasses both DANN and DAPT across all datasets, showing its effectiveness in adapting to domain shifts. Similarly, the buffered approach \onlinebuf demonstrates notable performance, outperforming both methods most of the time and consistently outperforms the baseline \tunesource. Between the cumulative approach \onlinecum and buffered approach \onlinebuf, \onlinecum consistently outperforms \onlinebuf. This highlights the advantage of retaining historical  examples in the buffer, provided there is sufficient time and memory capacity due to its higher time and space complexity.

An intriguing observation is the poor performance of \offline, despite being trained on a larger training set size compared to cumulative \onlinecum when $t < T$. Its performance is even worse than the \tunesource baseline. This shows the importance of accounting for gradual domain shifts in model training. Unlike \onlinecum and \onlinebuf, which processes the target domain one by one incrementally, \offline's inability to adapt gradually to evolving domains leads to its poor performance. The trend we observed above is similar across BERT and FLAN-T5-XXL.

\paragraph{Cumulative self-training is robust against evolving domain shift}

\begin{figure*}[htbp!] 
\centering
\begin{subfigure}{0.45\textwidth} 
\centering
\includegraphics[width=\linewidth]{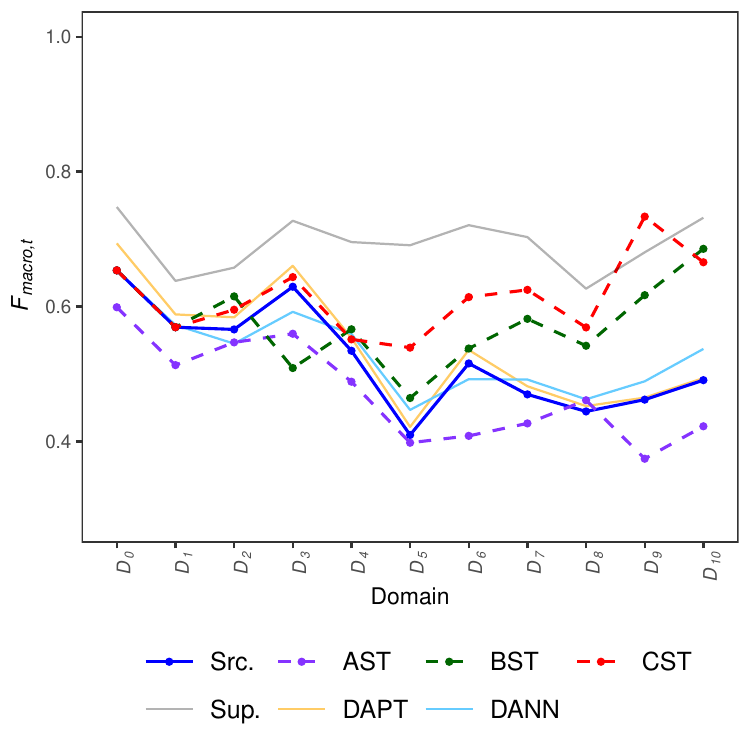}
\caption{COVID Dataset} 
\label{fig:merged_tuning_gda_covid}
\end{subfigure}
\hfill
\begin{subfigure}{0.45\textwidth} 
\centering
\includegraphics[width=\linewidth]{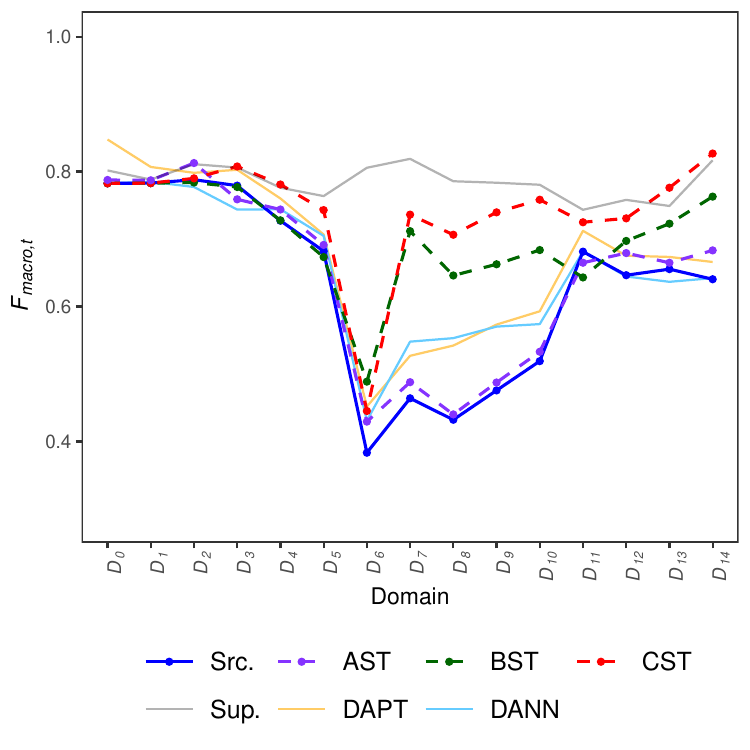}
\caption{WTWT Dataset} 
\label{fig:merged_tuning_gda_wtwt}
\end{subfigure}
\caption{Fine-tuning with self-training over time on both the (a) COVID and (b) WTWT dataset. The x-axis denotes the domains, beginning with the source domain and all subsequent target domains. The Macro-F1 score for each domain are plotted on the y-axis. Five different fine-tuning methods are represented in these plots: \tunesource (\textcolor{blue}{blue}), 
\offline (\textcolor{violet}{purple}),
\onlinebuf (\textcolor{teal}{green}), \onlinecum (\textcolor{red}{red}), and \tunetrue (\textcolor{darkgray}{grey}).} 
\label{fig:merged_tuning_gda_covid_wtwt}
\end{figure*}

Figures~\ref{fig:merged_tuning_gda_covid_wtwt} visualizes the model performance over domains for the COVID and the WTWT datasets. The overall trend for the PUBCLS and SCIERC dataset is similar and is shown Figure~\ref{fig:merged_tuning_gda_pubcls_scierc} in Appendix~\ref{sec:appendix-result-wtwt}. Firstly, the \tunesource baseline model, while yielding satisfactory performance in the source domain, encounters a gradual degradation over time. On the contrary, the cumulative approach \onlinecum exhibits remarkable resilience to changes over time across all datasets. The fixed-size buffered approach \onlinebuf also consistently outperforms the \tunesource baseline.

\label{sec:result_over_domain}

Additionally, Figure \ref{fig:merged_tuning_gda_covid_wtwt} highlights a noteworthy phenomenon. In the WTWT dataset, When data from domain $\Dcal_6$ is introduced, both \onlinebuf and \onlinecum undergo a sharp performance drop, mirroring the decline in \tunesource. However, unlike the baseline model, \onlinebuf and \onlinecum recover quickly in the subsequent domain domains. The key to this recovery lies in their ability to update the buffer with pseudolabeled examples from the $\Dcal_6$. As a result, the models is able to adapt to the domain shift.

\paragraph{Analysis of Evolving Domain Shift}

\begin{figure}[tbp!] 
\centering
\includegraphics[width=\linewidth]{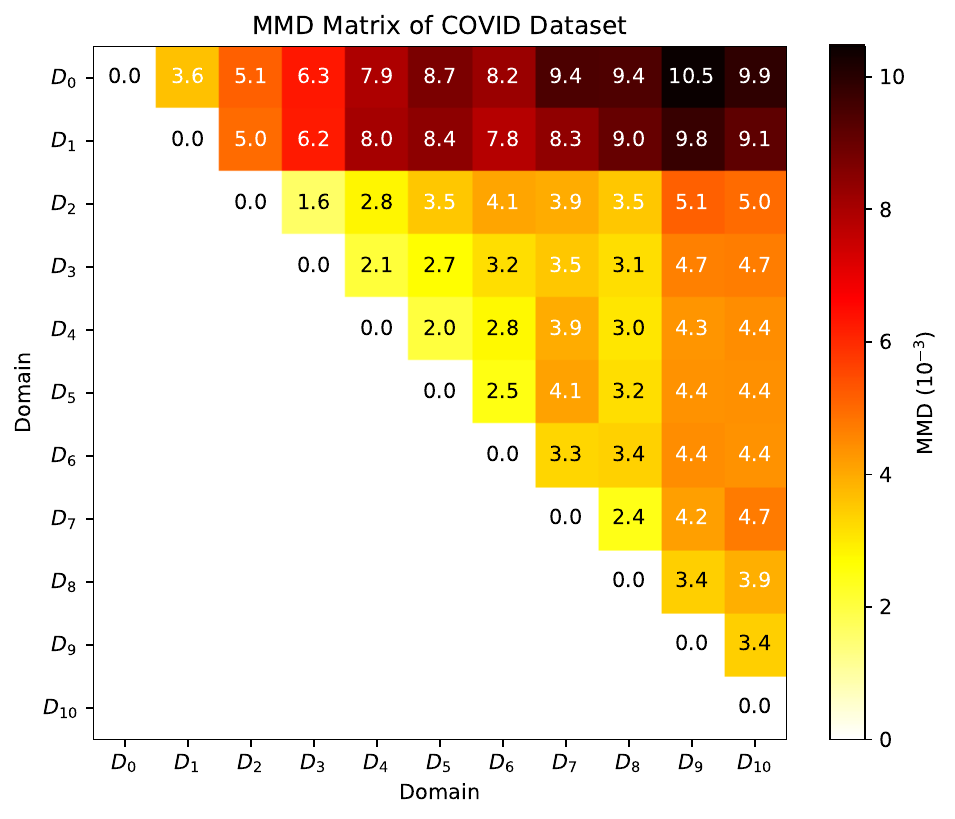}
\vspace{-4mm}
\caption{Maximum Mean Discrepancy (MMD) matrix of the COVID Dataset. Each cell represents the MMD between a pair of domains, calculated based on the marginal distribution of text embeddings $P_{g(\Xcal)}$ projected by the E5-Large-V2 model. The numbers in the heatmap are in the units of $10^{-3}$.}
\label{fig:mat_dist_mmd_covid_marginal}
\vspace{-5mm}
\end{figure}

To further understand the nature of evolving domain shift in our datasets, we visualize the covariate shift between its domains. We measure the Maximum Mean Discrepancy (MMD)\footnote{We use the RBF kernel for MMD.}\cite{gretton2012kernel} between the model embeddings of two domains using the E5-Large-V2 model \cite{wang2022text} to generate data embeddings. Figure~\ref{fig:mat_dist_mmd_covid_marginal} illustrates the MMD matrix for the COVID dataset, highlighting the differences between all pairs of domains within the dataset. A critical observation is that the MMD between two adjacent domains is almost always smaller than the MMD between the source and any target domain. The reduced MMD between adjacent domains justifies the incremental adaptation strategy like \onlinecum and \onlinebuf, as they can continually adjust to gradual domain changes, as opposed to one-pass \offline. This trend of smaller MMD between adjacent domains compared to those further apart is consistent across all four datasets we analyzed. For MMD matrices of the other datasets, please refer to Appendix~\ref{app:mmd_matrix_others}.

\begin{figure}[tbp!] 
\centering
\includegraphics[width=1.1\linewidth]{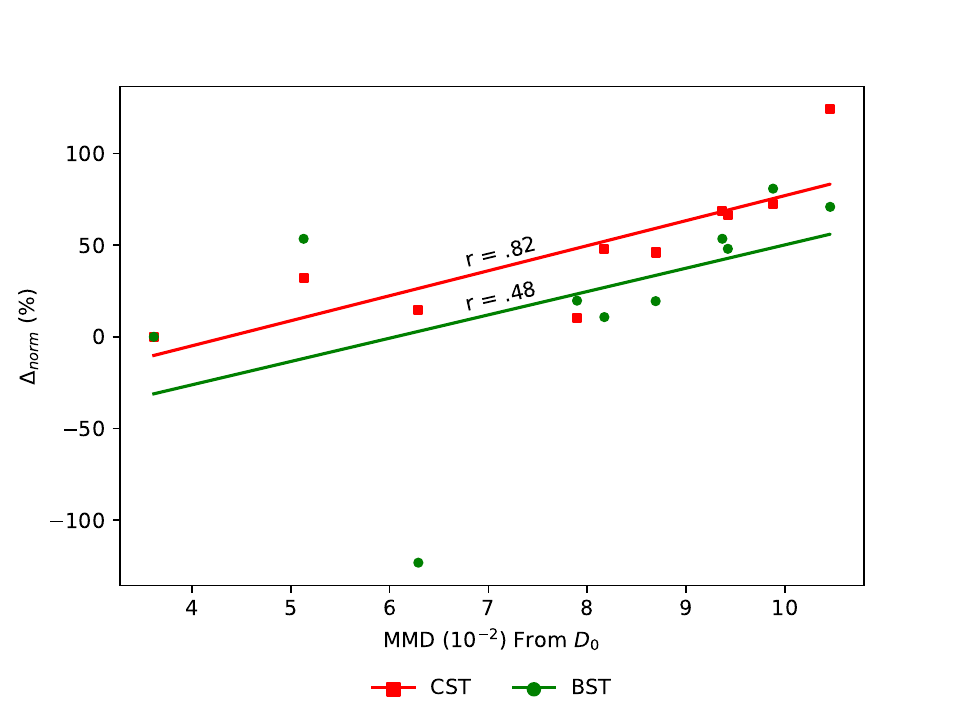}
\vspace{-4mm}
\caption{Scatter plot depicting, within the COVID dataset, the correlation between MMD (Maximum Mean Discrepancy) from the source domain $\Dcal_{0}$ to each target domain (X-axis) and the relative gain $\Delta_{\text{norm}}$ (\%) (Y-axis) for each target domain. Each dot represents a target domain, with \textcolor{red}{red} dots denoting results from \onlinecum and \textcolor{teal}{green} dots from \onlinebuf. The linear regression lines for both methods are overlaid, along with the Pearson's $r$ to indicate the strength of the relationship.}
\label{fig:scatter_plot_covid}
\vspace{-5mm}
\end{figure}

\paragraph{Benefits of \onlinebuf and \onlinecum Correlates with Domain Shift}
Figure~\ref{fig:scatter_plot_covid} demonstrates, within the COVID dataset, the correlation between the performance of \onlinebuf and \onlinecum in each target domain, and the degree of domain shift, as measured by Maximum Mean Discrepancy (MMD) from the source domain $\Dcal_{0}$. Each dot in the scatter plot represents a target domain, with red and green dots indicating the performance of \onlinecum and \onlinebuf methods, respectively. The plot reveals that relative gains increase with larger domain shifts, as shown by the regression lines and the Pearson's correlation coefficient. This trend underscores the adaptability of \onlinebuf and \onlinecum to evolving domain shifts, particularly when the divergence from the source domain is large. For the correlations in other datasets, please refer to Appendix \Scref{app:scatter_plot_others}, where this positive trend is similarly observed.

\paragraph{Ablation Study: Upsampling in Self-training is Critical}
\begin{table}[tbhp!]
    \centering
    \resizebox{\linewidth}{!}{
        \begin{tabular}{llllll}
            \toprule
            Setting & Method & \multicolumn{4}{c}{Dataset} \\
            \cmidrule{3-6}
            & & COVID & WTWT & SCIERC & PUBCLS \\
            \midrule
            Baseline & \tunesource & 0.509 & 0.618 & 0.485 & 0.406 \\
            \midrule                   
            Upsampling & \onlinebuf & 0.569(34$\uparrow$) & 0.697(47$\uparrow$) & 0.520(10$\uparrow$) & 0.469(26$\uparrow$) \\
            No Upsampling & \onlinebuf & 0.507(1$\downarrow$) & 0.703(50$\uparrow$) & 0.279(61$\downarrow$) & 0.372(13$\downarrow$) \\
            \midrule
            Upsampling & \onlinecum & 0.611(57$\uparrow$) & 0.739(72$\uparrow$) & 0.684(59$\uparrow$) & 0.492(35$\uparrow$) \\
            No Upsampling & \onlinecum & 0.511(1$\uparrow$) & 0.742(74$\uparrow$) & 0.461(7$\downarrow$) & 0.377(11$\downarrow$) \\
            \midrule            
            \tunetrue & - & 0.687 & 0.785 & 0.822 & 0.651 \\        
            \bottomrule
        \end{tabular}
    }
    \caption{Ablation study on the impact of upsampling in self-training methods. Performance metrics ($F_\text{avg}$) as well as $\Delta_{\text{avg,norm}}$ (in percentage in the parentheses (\%)) are shown for the \onlinebuf and \onlinecum methods, both with and without upsampling, across all datasets.}
    \label{tab:ablation_study_upsampling}
    \vspace{-5mm}
\end{table}

Given the presence of label shift in evolving domain data, our ablation study, presented in Table \ref{tab:ablation_study_upsampling}, is crucial to understand the impact of upsampling in self-training methods. The results show a significant reduction in the performance of \onlinebuf and \onlinecum when upsampling is omitted (WTWT is an exception with no significant difference). This consistent pattern across all datasets underlines the importance of upsampling, which ensures balanced class representation in each time step, thereby maintaining model effectiveness in dynamically changing environments. The decrease in performance without upsampling underscores its essential role in adapting to ongoing domain shifts.





\section{Related Work}
\label{sec:related}

\paragraph{Evolving Domain Adaptation} The \textit{Evolving Domain Adaptation} (EDA) or \textit{Continuous Domain Adaptation}~\cite{bobu2018adapting, wang2020continuously} problem requires continual adaptation of a model to data distributions that evolve over time~\cite{hoffman2014continuous,bitarafan2016incremental}.
This is closely related to the setting of \textit{Gradual Domain Adaptation} (GDA)~\cite{kumar2020understanding}, where target domain gradually shifts away from the source domain. Among many methods proposed for GDA, gradual self-training (GST) has been shown to be robust \cite{kumar2020understanding,wang2022understanding,chen2021gradual} when the domain shift is gradual. While GDA assumes the shift in conditional distribution $P(\mathcal{X}|\mathcal{Y})$ between adjacent domains to be small, EDA is not constraint by this assumption as the shift may be abrupt.

\paragraph{Text Classification under Evolving Domain Shift}
While existing works on EDA have been explored in multimodal data~\cite{wang2022understanding,chen2021gradual,kumar2020understanding,bitarafan2016incremental,hoffman2014continuous}, the problem has important implications in text Classification tasks like stance detection, entity recognition, among many. For example, online text data for stance detection have been evolving naturally over time through shifting opinions and narratives~\cite{alkhalifa2021opinions,alkhalifa2022capturing,mu2023examining}. \citet{alkhalifa2021opinions} propose to adapt word embeddings to tackle the stance detection task over time. However, to our best knowledge, a systematic investigation of various PLMs on text classification in the EDA setup is still missing.

\section{Conclusion}
We introduce two new evolving domain adaptation (EDA) methods that use dynamic buffering to mitigate the challenges posed by evolving domain shifts in text classification using PLMs. Our methods use fine-tuning of SLMs and prompting of LLMs. Our results highlight the importance of using up-to-date data for EDA, the significant role of intermediate domains, and the critical reliance of our strategies on accurate pseudo-labeling. Together, these insights offer an innovative perspective for addressing text classification in time-series data with pre-trained models.
\label{sec:conclusion}
\section*{Ethics Statement}

Our research introduces novel methods to enhance text classification performance under evolving domain shift, contributing to the reliability of language models that can have serious consequences if incorrectly predicted. Our approaches, by advancing the models' adaptability without requiring labeled data, can vastly benefit various applications, including assessing public opinion on critical issues like public health from social media.

Adhering to stringent ethical standards and legal compliance, this study anticipates no harmful outcomes. We utilize publicly available datasets, and we will release the de-identified data we collected in compliance with Twitter's policy. The annotation process, endorsed by our institute's Institutional Review Board (IRB), strictly follows all ethical guidelines. Upon study completion, we will release our code, promoting transparency and fostering further research into unsupervised domain adaptation in natural language processing.
\section*{Limitations}
First, the use of ChatGPT, a closed-source model, impedes understanding of underlying model mechanisms. Although we include FLAN-T5-XXL, an open-source model, to enhance our study, testing our methods on a broader spectrum of open-source models is warranted for a comprehensive understanding of evolving domain adaptation.

Second, our study only focus on stance detection tasks. There are other areas of text classification that may face similar EDS challenges, such as sentiment analysis and entity recognition. Future research should thus seek to evaluate our methods to these tasks. ensuring their widespread applicability.

Additionally, recent research suggests that chain-of-thought reasoning, augmented by labeled examples, can reach strong performance on stance detection \cite{zhang2023investigating}. Given this, it is worth investigating whether using up-to-date labeled examples can also aid this approach.


\bibliography{anthology,custom}

\newpage
\appendix
\label{sec:appendix}

\section{Result Figures on the Other datasets}
\label{sec:appendix-result-wtwt}
\subsection{Fine-tuning with Self-training}
\begin{figure*}[htbp!] 
\centering

\begin{subfigure}{0.48\textwidth} 
\centering
\includegraphics[width=\linewidth]{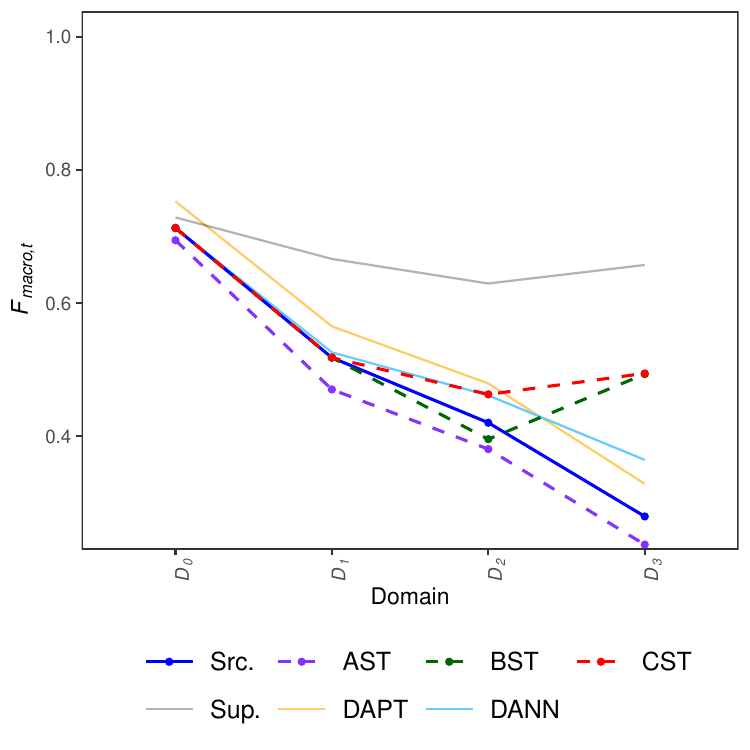}
\caption{PUBCLS Dataset} 
\label{fig:merged_tuning_gda_pubcls_2col}
\end{subfigure}
\hfill
\begin{subfigure}{0.48\textwidth} 
\centering
a\includegraphics[width=\linewidth]{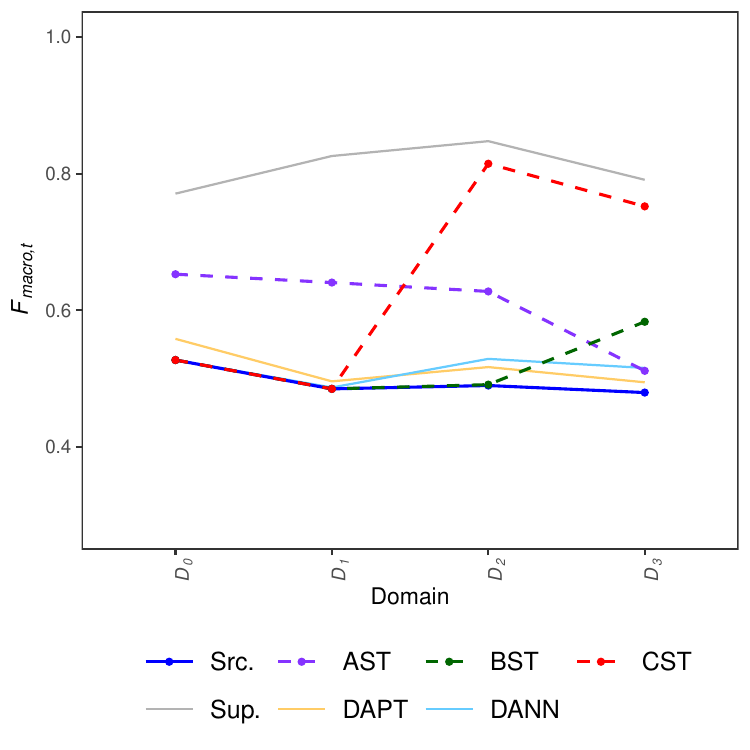}
\caption{SCIERC Dataset} 
\label{fig:merged_tuning_gda_scierc_2col}
\end{subfigure}

\caption{Fine-tuning with self-training over time on the (a) PUBCLS and (b) SCIERC datasets. The x-axis denotes the domains, beginning with the source domain and all subsequent target domains. The Macro-F1 score for each domain are plotted on the y-axis. Five different fine-tuning methods are represented in these plots: \tunesource (\textcolor{blue}{blue}), \onlinebuf (\textcolor{teal}{green}), \onlinecum (\textcolor{red}{red}), \offline (\textcolor{violet}{purple}), and \tunetrue (\textcolor{darkgray}{grey}).}
\label{fig:merged_tuning_gda_pubcls_scierc}
\end{figure*}

Figure~\ref{fig:merged_tuning_gda_pubcls_scierc} visualizes the model performance over domains for the PUBCLS and SCIERC dataset when using fine-tuning with self-training.

\subsection{Analysis of Evolving Domain Shift}\label{app:mmd_matrix_others}

\begin{figure}[htbp!] 
\centering
\includegraphics[width=\linewidth]{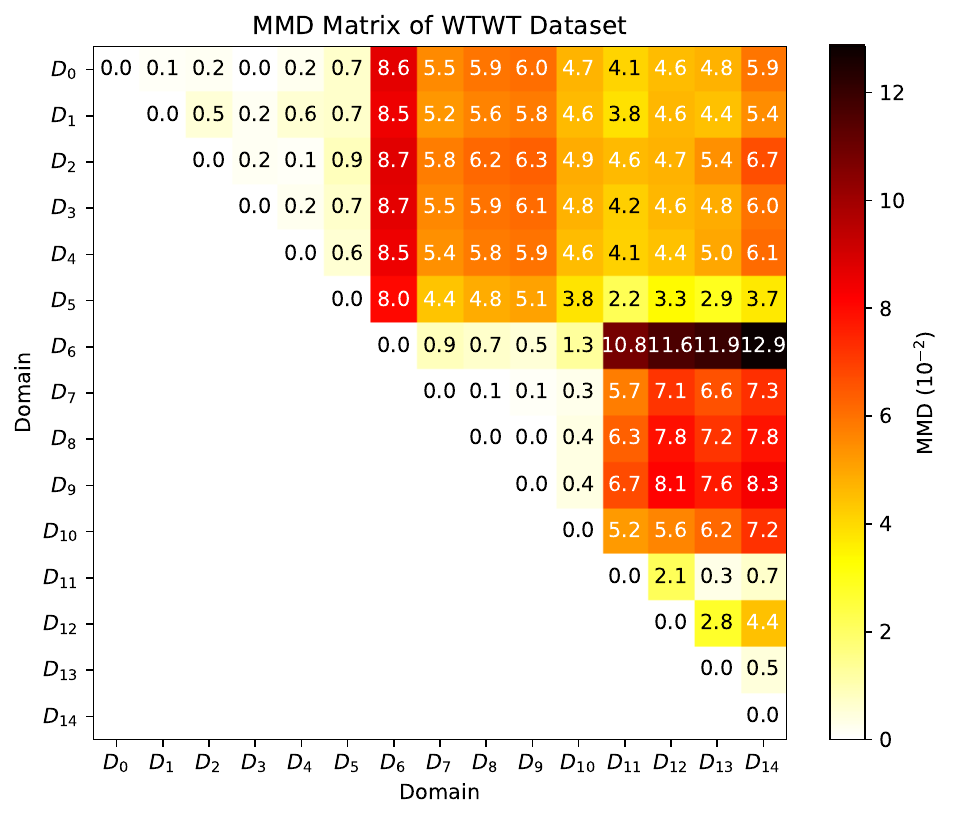}
\caption{Maximum Mean Discrepancy (MMD) matrix of the WTWT Dataset. Each cell represents the MMD between a pair of domains, calculated based on the marginal distribution of text embeddings $P_{g(\Xcal)}$ projected by the E5-Large-V2 model. The color gradient ranges from white (representing zero discrepancy) to darker shades of red (indicating larger discrepancies). Please note that the numbers in the heatmap are in the units of $10^{-2}$.}
\label{fig:mat_dist_mmd_wtwt_marginal}
\end{figure}

\begin{figure}[htbp!] 
\centering
\includegraphics[width=\linewidth]{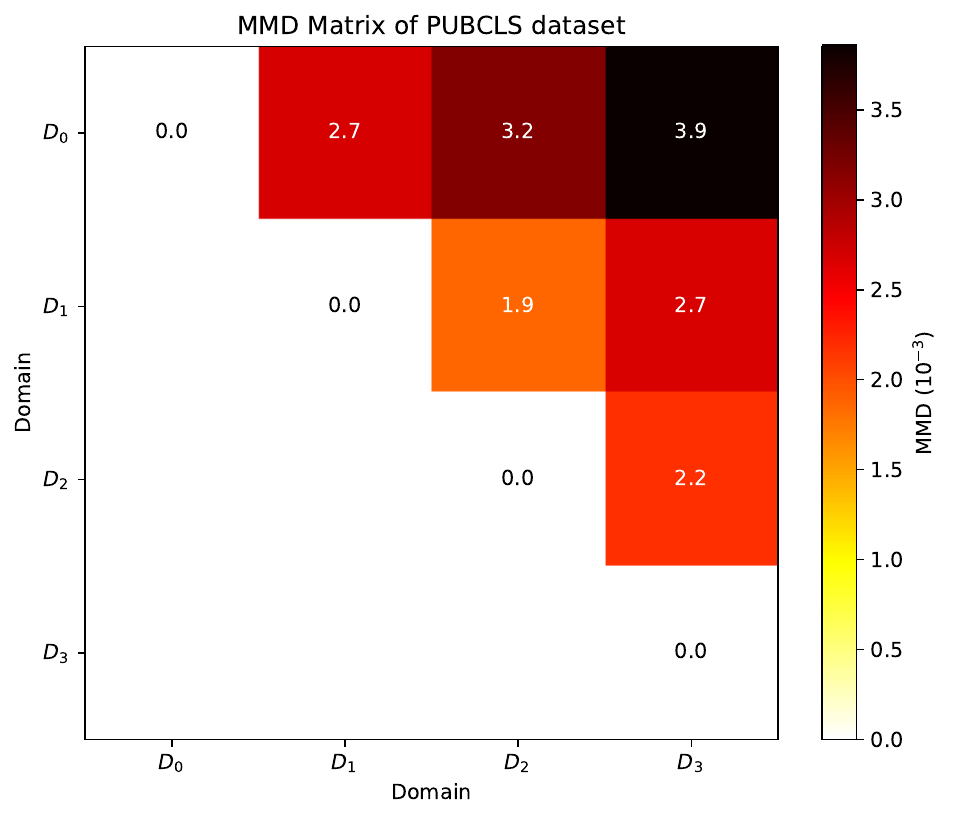}
\caption{Maximum Mean Discrepancy (MMD) matrix of the PUBCLS Dataset. Each cell represents the MMD between a pair of domains, calculated based on the marginal distribution of text embeddings $P_{g(\Xcal)}$ projected by the E5-Large-V2 model. The color gradient ranges from white (representing zero discrepancy) to darker shades of red (indicating larger discrepancies). Please note that the numbers in the heatmap are in the units of $10^{-3}$.}
\label{fig:mat_dist_mmd_pubcls_marginal}
\end{figure}

\begin{figure}[htbp!] 
\centering
\includegraphics[width=\linewidth]{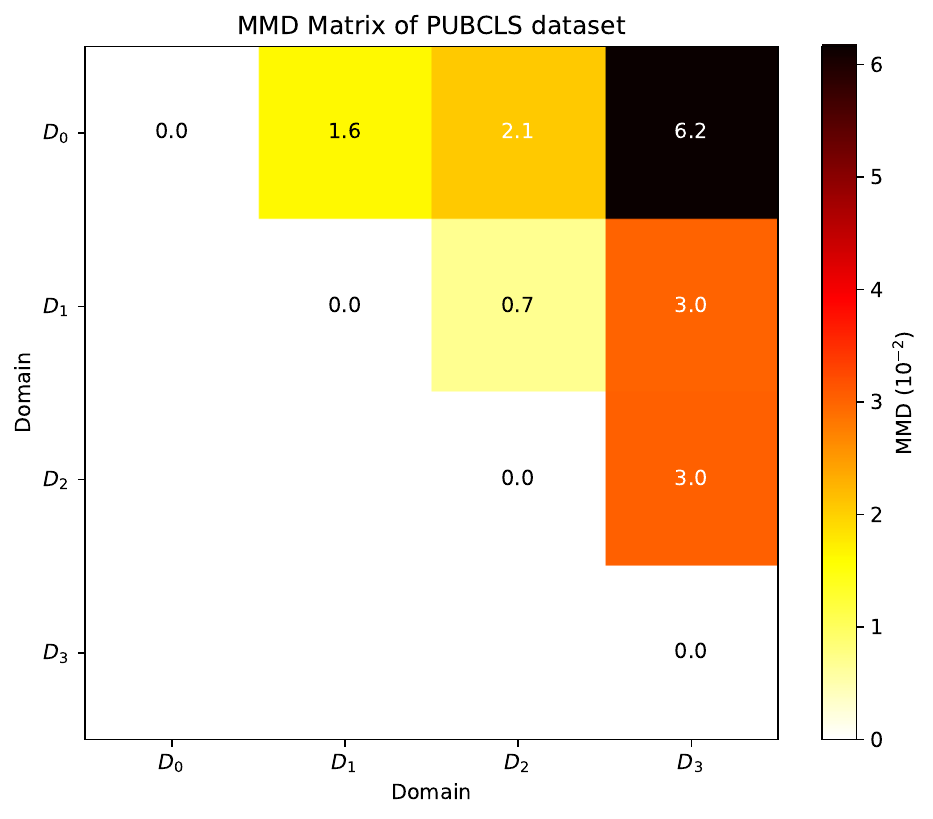}
\caption{Maximum Mean Discrepancy (MMD) matrix of the SciERC Dataset. Each cell represents the MMD between a pair of domains, calculated based on the marginal distribution of text embeddings $P_{g(\Xcal)}$ projected by the E5-Large-V2 model. The color gradient ranges from white (representing zero discrepancy) to darker shades of red (indicating larger discrepancies). Please note that the numbers in the heatmap are in the units of $10^{-2}$.}
\label{fig:mat_dist_mmd_sci_erc_marginal}
\end{figure}

This appendix extends our evolving domain shift analysis to the WTWT, PUBCLS, and SCIERC datasets, complementing the COVID dataset findings (Figure~\ref{fig:mat_dist_mmd_covid_marginal}). Using Maximum Mean Discrepancy (MMD), we visualize covariate shifts within these datasets (Figure~\ref{fig:mat_dist_mmd_wtwt_marginal},\ref{fig:mat_dist_mmd_pubcls_marginal},\ref{fig:mat_dist_mmd_sci_erc_marginal}), highlighting that the MMD between adjacent domains tends to be smaller than that between the source and target domains. This pattern is true across all datasets, underscoring the benefits of iterative self-training methods (\onlinecum and \onlinebuf) over single-pass approaches like \offline.

\subsection{Benefits of \onlinebuf and \onlinecum Correlates with Domain Shift}\label{app:scatter_plot_others}

\begin{figure}[tbp!]
    \centering
    
\begin{subfigure}[b]{\linewidth}
        \centering
        \includegraphics[width=\linewidth]{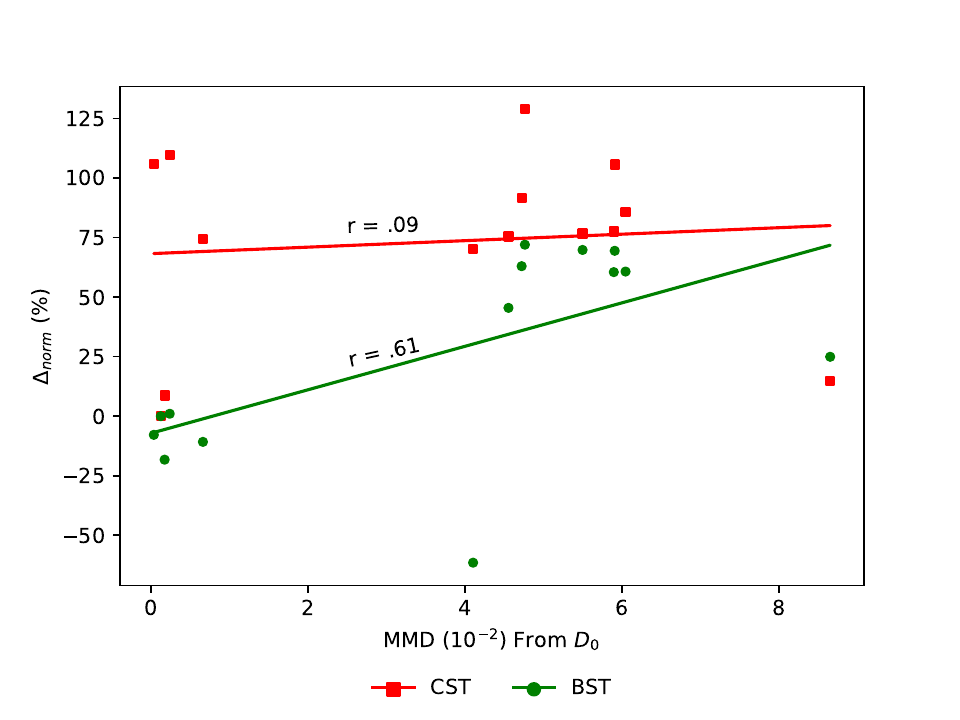}
        \caption{WTWT dataset}
        \label{fig:scatter_plot_wtwt}
    \end{subfigure}
    \vspace{2mm} 

    \begin{subfigure}[b]{\linewidth}
        \centering
        \includegraphics[width=\linewidth]{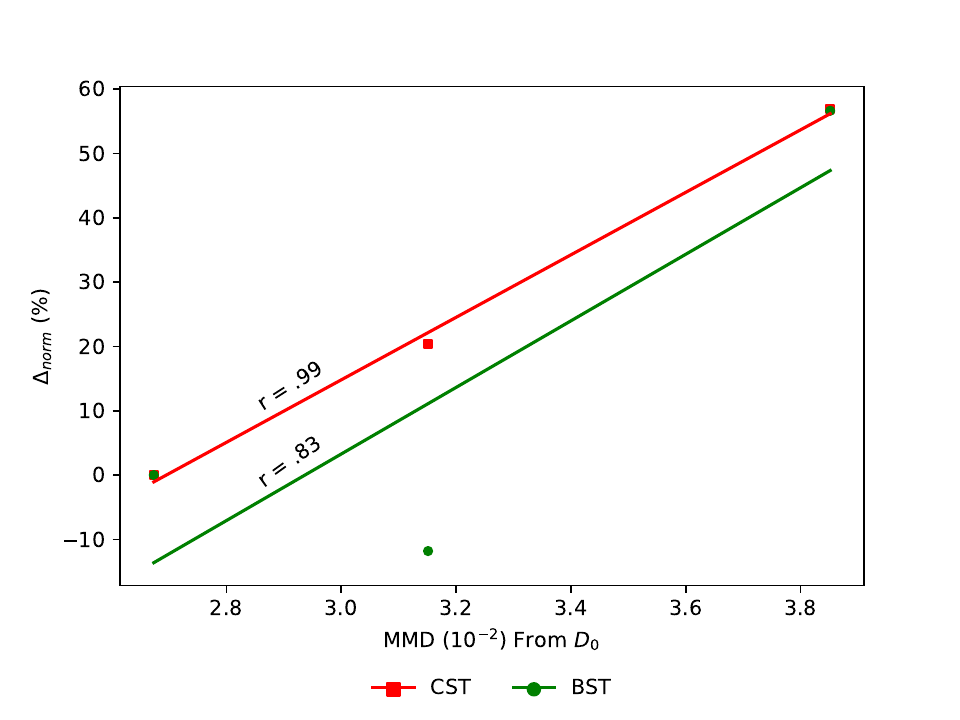}
        \caption{PUBCLS dataset}
        \label{fig:scatter_plot_pubcls}
    \end{subfigure}
    \vspace{2mm} 

    \begin{subfigure}[b]{\linewidth}
        \centering
        \includegraphics[width=\linewidth]{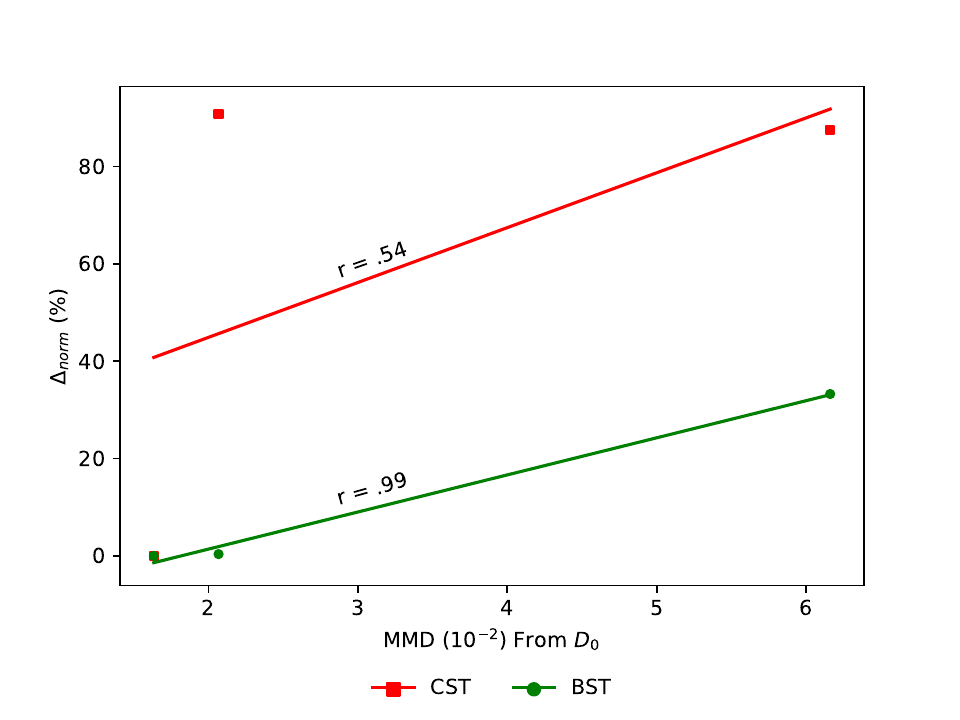}
        \caption{SCIERC dataset}
        \label{fig:scatter_plot_scierc}
    \end{subfigure}

    \caption{Scatter plots depicting the correlation between MMD (Maximum Mean Discrepancy) from the source domain $\Dcal_{0}$ to each target domain (X-axis) and the relative gain $\Delta_{\text{norm}}$ (\%) (Y-axis) for each target domain in the (a) WTWT, (b) PUBCLS, and (c) SCIERC datasets. Each dot represents a target domain, with \textcolor{red}{red} dots denoting results from \onlinecum and \textcolor{teal}{green} dots from \onlinebuf. Linear regression lines for both methods are overlaid, along with the Pearson's $r$ to indicate the strength of the relationship.}
    \label{fig:scatter_plots_wtwt_pubcls_scierc}
\end{figure}

Figure~\ref{fig:scatter_plots_wtwt_pubcls_scierc} visualizes scatter plots for the WTWT, PUBCLS, and SCIERC datasets, respectively, each showing the positive correlation between the performance of \onlinebuf and \onlinecum and domain shift, similar to the trend in the COVID dataset (Figure~\ref{fig:scatter_plot_covid}).

\section{COVID-19 Vaccination (COVID) Dataset}
\label{sec:appendix-covid}

\subsection{Data Retrieval}

We collected Twitter data via Twitter Academic API 2.0 endpoint using a list of keywords related to COVID-19 vaccination in English (see Table 6 for details). The time frame of the dataset is from December 1, 2020 to June 30, 2022. We took a random sample on daily basis for human annotation in terms of valence classification (\textit{N} = 9,373) into two levels, “against” or “not-against” COVID-19 vaccination.
\subsection{Task Definition and Annotation Guidelines}
The main goal of the human annotation is to identify the valence toward COVID-19 vaccination of each tweet. We classified the valence into two categories, which included “against” and “non-against” labels in terms of COVID-19 vaccination:

1.	The “Against” label can be a) the author of the tweet is personally against COVID-19 vaccines (anti-vaccine) or vaccination policies; b) the tweet message indicates negative consequences of COVID-19 vaccination, such as severe side effects or health misinformation; etc.

E.g., “\textit{Pfizer’s Covid Jabs Shown to Decrease Male Fertility for Months After Vaccination}” and “\textit{My son-in-law committed suicide today. He was vaxxed, boosted x 2.  Started losing weight and lost control of his bladder, had to be catheterized. He weighed 149 lbs at his death. Tests were pending to see what was wrong. He left a 11 yr old son and 5 yr old daughter.}”

2. The “Not-against” label can be a) the author of the tweet personally supports or promotes COVID-19 vaccines (pro-vaccine); b) the tweet message reports positive news of COVID-19 vaccines; or c) the tweet is ambiguous to identify its valence.

E.g., “\textit{Good morning. Please get vaccinated}” and “\textit{By the way, vaccination is not a ‘deeply personal decision.’ It is a routine public health requirement in a civilized society.}”  

\subsection{Data Annotation}
Eight volunteers pursuing undergraduate studies were recruited to annotate the Twitter data, with each tweet being annotated by three different annotators. Prior to the annotation task, the annotators underwent a comprehensive training process. The annotation process took place over nine months, from December 2021 to August 2022.

The annotation task was divided into two main steps: 1) Relevancy: Annotators first determined whether each tweet was relevant to the subject of COVID-19 vaccination. This acted as a screening question to filter out unrelated content. 2) Stance: If a tweet was deemed relevant, the annotators were required to assign it an "against" or "non-against" label. Out of the original 9,373 tweets, 5,002 were considered relevant to COVID-19 vaccination. In instances where the three annotators disagreed on the coding of a tweet, a majority vote rule was applied to reach a final decision. This rule was chosen due to its efficacy in resolving disagreements in human annotations.

\subsection{Quality Assessment}
Given that each tweet was annotated by three annotators from a team of eight, we calculated the inter-coder reliabilities for each three-person sub-team (see Table~\ref{tab:covid_krippendorffs_alphas} for details). The resulting weighted-average Krippendorff's alpha, weighted by the number of samples annotated by each sub-team, was 0.64. This Krippendorff's alpha is deemed acceptable as it exceeds 0.6 \cite{wong-etal-2021-cross,landis1977measurement}.

To evaluate the quality of the majority vote rule, one of the authors, an expert in this field, randomly selected 300 tweets and provided expert annotations as gold labels. The comparison between the annotators' labels and the gold labels resulted in an accuracy (percentage agreement) of 89.7\%  indicating a high level of concordance and thereby affirming the reliability of our annotation process (Table~\ref{tab:covid_confusion_matrix} shows the agreement matrix).

\begin{table}[htbp!]
\centering
\small
\begin{tabularx}{\columnwidth}{X}
\toprule
vaccine, vaccines, vaccination, vaccinations, vaccinate, vaccinated, vax, vaxx, vaxxx, vaxxed, covax, shot, shots, dose, doses, covidvaccine, covid19vaccine, coronavaccine, coronavirusvaccine, covaxin, mrna, nvic, booster, boosters, pfizer, moderna, gamaleya, “oxford-astrazeneca”, astrazeneca, cansino, “johnson \& johnson”, “j\&j”, “j \& j”, “vector institute”, novavax, sinopharm, sinovac, “bharat biotech”, janssen, cepi, biontech, sputnikv, bektop, zfsw, nvic, pfizerbiontech, “biontechvaccine”, “warp speed”, “delta variant”, oxfordvaccine, pfizervaccine, pfizercovidvaccine, modernavaccine, modernacovidvaccine, biotechvaccine, biotechcovidvaccine, biontechvaccine, biontechcovidvaccine, bektopvaccine, simopharmvaccine, johnsonvaccine, janssenvaccine, azvaccine, astrazenecacovidvaccine, astrazenecavaccine, thisisourshot, vaxhole, notocoronavirusvaccines, getvaccinated\\
\bottomrule
\end{tabularx}
\caption{The keyword list for COVID-19 vaccine Twitter data collection}
\label{tab:keywords_covid}
\end{table}
\begin{table}[htbp!]
\centering
\small
\begin{tabularx}{\columnwidth}{lll}
\toprule
Sub-team Index & Percentage of Tweets (\%) & $\alpha$ \\
\midrule
 1 & 3\% & 0.56 \\
 2 & 14\% & 0.56 \\
 3 & 11\% & 0.64 \\
 4 & 33\% & 0.64 \\
 5 & 15\% & 0.70 \\
 6 & 5\% & 0.61 \\
 7 & 4\% & 0.55 \\
 8 & 14\% & 0.69 \\
\midrule
Weighted $\alpha$ & 100\% & 0.64 \\
\bottomrule
\end{tabularx}
\caption{Krippendorff’s alphas ($\alpha$) for annotator teams}
\label{tab:covid_krippendorffs_alphas}
\end{table}
\begin{table}[htbp!]
\centering
\small
\begin{tabular}{llcc}
\toprule
& & \multicolumn{2}{c}{Expert Gold Label} \\
\cmidrule(lr){3-4}
& & Not Against & Against \\
\midrule
\multirow{2}{*}{Annotators' Label} & Not Against & 193 & 8 \\
& Against & 23 & 76 \\
\bottomrule
\end{tabular}
\caption{Agreement Matrix: Annotations' Label vs Expert Gold Label}
\label{tab:covid_confusion_matrix}
\end{table}





\section{Data Preprocessing}
\label{sec:data-preprocess}

\subsection{Partitioning the Data Chronologically}

To study Evolving Domain Adaptation (EDA), we partitioned the dataset in a chronological manner. This approach ensured the oldest instances served as the labeled source domain, while subsequent instances were arranged into a series of target domains based on their timestamps. This alignment, by natural time units like months instead of fixed-size partitions, is designed to emulate real-world scenarios. For the COVID dataset, we used one-month units for partitioning, and for the WTWT dataset, we used two-month units, with a few exceptions, e.g., period from February to March 2022 was mergd into a single domain to ensure an adequate number of instances for reliable evaluation. 

\subsection{Source and Target Domains of the COVID-19 Vaccination Dataset (COVID)}

For the COVID dataset, instances ranging from December 2020 to May 2021 were merged to form the source domain. The rest of the instances from June 2021 to June 2022 were used to create 10 target domains using a one-month interval. For the detailed correspondence between times and domain, please refer to Table~\ref{tab:label_dist_over_time_covid}.

\subsection{Source and Target Domains of the Will-They-Won’t-They Dataset (WTWT)}

For the WTWT dataset, instances from June 2015 to June 2016 were combined to create the source domain. The remaining instances from July 2016 to December 2018 were used to create 14 target domains using a two-month interval. For the detailed correspondence between times and domain, please refer to Table~\ref{tab:label_dist_over_time_wtwt}.

\subsection{Training, Validation, Testing Partitions}
\label{sec:appendix-train_vali_partition}

Each domain was further divided into training, validation, and test sets in a 5:1:4 ratio. The testing set was not used during the training process, and is only used to evaluate the model performance at each domain. Only instances from the training and validation sets were pseudolabeled if the method required pseudolabeling. The validation set was used for fine-tuning, allowing us to select the best epoch checkpoint (see Appendix~\ref{sec:appendix-tuning_hyperparams}).

\section{Label and Topic Distribution over Domains}
\label{sec:appendix-dist-label-topic}

Table~\ref{tab:label_dist_over_time_covid}, \ref{tab:label_dist_over_time_wtwt}, \ref{tab:topic_dist_over_time_wtwt},
\ref{tab:label_dist_over_time_pubcls}, and 
\ref{tab:label_dist_over_time_sci_erc} present the label and topic distribution shifts across different domains. 
For example, for the COVID dataset (Table~\ref{tab:label_dist_over_time_covid}), the proportion of tweets against COVID-19 vaccines decreased over domains.



\begin{table*}[htbp!]
\centering
\small
\begin{tabular}{lllll}
\toprule
Domain Type & Domain & Against & Not-against & Total \\
& & Count (\%) & Count (\%) & \\
\midrule
Source Domain & $\Dcal_0$: 2020-12 to 2021-05 & 224 (14.34\%) & 1338 (85.66\%) & 1562 \\
\cmidrule{2-5}
Target Domains & $\Dcal_1$: 2021-06 & 51 (22.67\%) & 174 (77.33\%) & 225 \\
& $\Dcal_2$: 2021-07 & 109 (25.77\%) & 314 (74.23\%) & 423 \\
& $\Dcal_3$: 2021-08 & 132 (26.4\%) & 368 (73.6\%) & 500 \\
& $\Dcal_4$: 2021-09 & 160 (37.04\%) & 272 (62.96\%) & 432 \\
& $\Dcal_5$: 2021-10 & 157 (46.73\%) & 179 (53.27\%) & 336 \\
& $\Dcal_6$: 2021-11 & 117 (37.86\%) & 192 (62.14\%) & 309 \\
& $\Dcal_7$: 2021-12 & 142 (36.41\%) & 248 (63.59\%) & 390 \\
& $\Dcal_8$: 2022-01 & 144 (40.22\%) & 214 (59.78\%) & 358 \\
& $\Dcal_9$: 2022-02 to 2022-03 & 132 (56.65\%) & 101 (43.35\%) & 233 \\
& $\Dcal_{10}$: 2022-04 to 2022-06 & 133 (56.84\%) & 101 (43.16\%) & 234 \\
\midrule
Total & - &1501 (30.01\%) & 3501 (69.99\%) & 5002 \\
\bottomrule
\end{tabular}
\caption{Label Distribution of the COVID Dataset across Domains}
\label{tab:label_dist_over_time_covid}
\end{table*}

\begin{table*}[htbp!]
  \centering
  \small
  \resizebox{\linewidth}{!}{
\begin{tabular}{lllllll}
\toprule
Domain Type & Domain & Comment & Refute & Support & Unrelated & Total \\
& & Count (\%) & Count (\%) & Count (\%) & Count (\%) &     \\
\midrule
Source Domain & $\Dcal_0$: 2015-06 to 2016-06& 2923 (34.21 \%) & 1515 (17.72\%) & 1020 (11.93\%) & 3087 (36.14\%) & 8545 \\
\cmidrule{2-7}
Target Domains & $\Dcal_1$: 2016-07 to 2016-08 & 798 (36.44\%) & 331 (15.11\%) & 251 (11.46\%) & 810 (36.99\%) & 2190 \\
& $\Dcal_2$: 2016-09 to 2016-10 & 309 (34.22\%) & 148 (16.39\%) & 96 (10.63\%) & 350 (38.76\%) & 903 \\
& $\Dcal_3$: 2016-11 to 2016-12 & 307 (35.05\%) & 145 (16.55\%) & 99 (11.3\%) & 325 (37.1\%) & 876 \\
& $\Dcal_4$: 2017-01 to 2017-02 & 239 (30.6\%) & 120 (15.36\%) & 71 (9.09\%) & 351 (44.94\%) & 781 \\
& $\Dcal_5$: 2017-03 to 2017-06 & 212 (24.62\%) & 75 (8.71\%) & 64 (7.43\%) & 510 (59.23\%) & 861 \\
& $\Dcal_6$: 2017-07 to 2017-08 & 904 (45.75\%) & 37 (1.87\%) & 130 (6.58\%) & 905 (45.8\%) & 1976 \\
& $\Dcal_7$: 2017-09 to 2017-10 & 1247 (46.6\%) & 40 (1.49\%) & 332 (12.41\%) & 1057 (39.5\%) & 2676 \\
& $\Dcal_8$: 2017-11 to 2017-12 & 4782 (49.91\%) & 174 (1.82\%) & 1082 (11.29\%) & 3543 (36.98\%) & 9581 \\
& $\Dcal_9$: 2018-01 to 2018-02 & 2342 (47.05\%) & 136 (2.73\%) & 491 (9.86\%) & 2009 (40.36\%) & 4978 \\
& $\Dcal_{10}$: 2018-03 to 2018-04 & 2006 (46.86\%) & 141 (3.29\%) & 561 (13.1\%) & 1573 (36.74\%) & 4281 \\
& $\Dcal_{11}$: 2018-05 to 2018-06 & 454 (43.74\%) & 72 (6.94\%) & 177 (17.05\%) & 335 (32.27\%) & 1038 \\
& $\Dcal_{12}$: 2018-07 to 2018-08 & 367 (34.46\%) & 185 (17.37\%) & 311 (29.2\%) & 202 (18.97\%) & 1065 \\
& $\Dcal_{13}$: 2018-09 to 2018-10 & 523 (39.29\%) & 89 (6.69\%) & 398 (29.9\%) & 321 (24.12\%) & 1331 \\
& $\Dcal_{14}$: 2018-11 to 2018-12 & 447 (39.52\%) & 82 (7.25\%) & 323 (28.56\%) & 279 (24.67\%) & 1131 \\
\midrule
Total & - & 17860 (42.31\%) & 3290 (7.79\%) & 5406 (12.81\%) & 15657 (37.09\%) & 42213 \\
\bottomrule
\end{tabular}
}
\caption{Label Distribution of the WTWT Dataset across Domains}
  \label{tab:label_dist_over_time_wtwt}
\end{table*}

\begin{table*}[htbp!]
  \centering
    \small
    \resizebox{\linewidth}{!}{
    \begin{tabular}{llllllll}
    \toprule
        Domain Type & Domain & AET\_HUM & ANTM\_CI & CI\_ESRX & CVS\_AET & FOXA\_DIS  & Total \\
        & & Count (\%) & Count (\%) & Count (\%) & Count (\%) & Count (\%)  \\ 
        \midrule
        Source Domain & $\Dcal_0$: 2015-06 to 2016-06 & 3873 (45.32\%) & 4672 (54.68\%) & 0 (0\%) & 0 (0\%) & 0 (0\%) & 8545 \\
        \cmidrule{2-8}
        Target Domains & $\Dcal_1$: 2016-07 to 2016-08 & 1298 (59.27\%) & 890 (40.64\%) & 0 (0\%) & 0 (0\%) & 2 (0.09\%) & 2190 \\
        & $\Dcal_2$: 2016-09 to 2016-10 & 282 (31.23\%) & 621 (68.77\%) & 0 (0\%) & 0 (0\%) & 0 (0\%) & 903 \\
        & $\Dcal_3$: 2016-11 to 2016-12 & 377 (43.04\%) & 499 (56.96\%) & 0 (0\%) & 0 (0\%) & 0 (0\%) & 876 \\
        & $\Dcal_4$: 2017-01 to 2017-02 & 136 (17.41\%) & 550 (70.42\%) & 0 (0\%) & 95 (12.16\%) & 0 (0\%) & 781 \\
        & $\Dcal_5$: 2017-03 to 2017-06 & 0 (0\%) & 402 (46.69\%) & 30 (3.48\%) & 429 (49.83\%) & 0 (0\%) & 861 \\
        & $\Dcal_6$: 2017-07 to 2017-08 & 0 (0\%) & 0 (0\%) & 38 (1.92\%) & 117 (5.92\%) & 1821 (92.16\%) & 1976 \\
        & $\Dcal_7$: 2017-09 to 2017-10 & 0 (0\%) & 0 (0\%) & 100 (3.74\%) & 755 (28.21\%) & 1821 (68.05\%) & 2676 \\
        & $\Dcal_8$: 2017-11 to 2017-12 & 0 (0\%) & 0 (0\%) & 96 (1\%) & 2602 (27.16\%) & 6883 (71.84\%) & 9581 \\
        & $\Dcal_9$: 2018-01 to 2018-02 & 0 (0\%) & 0 (0\%) & 56 (1.12\%) & 1230 (24.71\%) & 3692 (74.17\%) & 4978 \\
        & $\Dcal_{10}$: 2018-03 to 2018-04 & 2 (0.05\%) & 0 (0\%) & 852 (19.9\%) & 843 (19.69\%) & 2584 (60.36\%) & 4281 \\
        & $\Dcal_{11}$: 2018-05 to 2018-06 & 0 (0\%) & 0 (0\%) & 185 (17.82\%) & 853 (82.18\%) & 0 (0\%) & 1038 \\
        & $\Dcal_{12}$: 2018-07 to 2018-08 & 0 (0\%) & 0 (0\%) & 624 (58.59\%) & 441 (41.41\%) & 0 (0\%) & 1065 \\
        & $\Dcal_{13}$: 2018-09 to 2018-10 & 0 (0\%) & 0 (0\%) & 176 (13.22\%) & 1155 (86.78\%) & 0 (0\%) & 1331 \\
        & $\Dcal_{14}$: 2018-11 to 2018-12 & 1 (0.09\%) & 0 (0\%) & 0 (0\%) & 1130 (99.91\%) & 0 (0\%) & 1131 \\
        \midrule
        Total & - & 5969 (14.14\%) & 7634 (18.08\%) & 2157 (5.11\%) & 9650 (22.86\%) & 16803 (39.81\%) & 42213 \\ 
        \bottomrule
    \end{tabular}
    }
    \caption{Topic Distribution of the WTWT Dataset across Domains} 
  \label{tab:topic_dist_over_time_wtwt}
\end{table*}

\begin{table*}[htbp!]
  \centering
  \small
  \resizebox{\linewidth}{!}{
        \begin{tabular}{llllll}
        \toprule
        Domain Type & Domain & Fox News & New York Times & Washington Post & Total  \\
        & & Count (\%) & Count (\%) & Count (\%) & Count (\%)      \\
        \midrule
        Source Domain & $\Dcal_0$: 2009 to 2010& 276 (17.25 \%) & 921 (57.56 \%) & 403 (25.19\%) & 1600 \\
        \cmidrule{2-6}
        Target Domains & $\Dcal_1$: 2011 to 2012 & 533 (33.31\%) & 748 (47.65\%) & 319 (19.94\%) & 1600 \\
        & $\Dcal_2$: 2013 to 2014 & 482 (30.12\%) & 544 (34.00\%) & 574 (35.88\%) & 1600 \\
        & $\Dcal_3$: 2015 to 2016 & 321 (20.06\%) & 758 (47.38\%) & 521 (32.56\%) & 1600 \\
        
        \midrule
        Total & - & 1612 (25.19\%) & 2971 (46.42\%) & 1817 (28.39\%) & 6400 \\
        \bottomrule
        \end{tabular}
}
\caption{Label Distribution of the PUBCLS Dataset across Domains}
  \label{tab:label_dist_over_time_pubcls}
\end{table*}

\begin{table*}[htbp!]
  \centering
  \small
  \resizebox{\linewidth}{!}{
        \begin{tabular}{lllllllll}
        \toprule
        Domain Type & Domain & Generic & Material & Method & Metric & Other Scientific Term & Task &Total  \\
        & & Count (\%) & Count (\%) & Count (\%) & Count (\%) &  Count (\%) &    \\
        \midrule
        Source Domain & $\Dcal_0$: 1980 to 1999 & 337 (17.05 \%) & 157 (7.95 \%) & 563 (28.49\%) & 37(1.87\%) & 555(28.09\%) & 327(16.55\%) & 1976 \\
        \cmidrule{2-9}
        Target Domains & $\Dcal_1$: 2000 to 2004 & 293 (15.64\%) & 186 (9.93\%) & 540 (28.82\%) & 84(4.48\%) & 478(25.51\%) & 293(15.64\%) & 1874\\
        & $\Dcal_2$: 2005 to 2009 & 372 (16.24\%) & 253 (11.05\%) & 548 (23.93\%) & 136(5.94\%) & 616(26.9\%) & 365(15.94\%) & 2290\\
        & $\Dcal_3$: 2010 to 2016 & 335 (17.14\%) & 175 (8.96\%) & 443 (22.67\%) & 82(4.2\%) & 621(31.78\%) & 298(15.25\%) & 1954\\
        
        \midrule
        Total & - & 1337 (16.52\%) & 771 (9.53\%) & 2094 (25.87\%) & 339(4.19\%) & 2270(28.05\%) & 1283(15.85\%) & 8094 \\
        \bottomrule
        \end{tabular}
        }
\caption{Label Distribution of the \text{SCI\_ERC} Dataset across Domains}
  \label{tab:label_dist_over_time_sci_erc}
\end{table*}


\section{Prompt Tempate}
\label{sec:appendix-prompt-template}
Table~\ref{tab:prompt-template-zero} shows the prompts used in the experiment.

\begin{table*}[htbp!]
  \centering
  \small
\begin{tabularx}{\textwidth}{ccX}
\toprule
Dataset & Components & Contents \\
\midrule
COVID & \textcolor{blue}{$x_i$} & \textcolor{blue}{breaking report: cdc used rejected study from india on vaccine, not approved in the us to justify new mask mandate...}\\\\
    & $\Tcal($\textcolor{blue}{$x_i$}$)$ & What is the stance of the tweet below with respect to COVID-19 vaccine? Please use exactly one word from the following 2 categories to label it: ‘against’, ‘not-against’. Here is the tweet. ‘\textcolor{blue}{breaking report: cdc used rejected study from india on vaccine, not approved in the us to justify new mask mandate...}’ The stance of the tweet is\\
\midrule
WTWT & \textcolor{blue}{$x_i$} & \textcolor{blue}{feed time... health ins. aetna to acquire humana for 37, anthem and cigna are in talks, centene corp. agreeing to buy health net \#uniteblue} \\\\
    & $\Tcal($\textcolor{blue}{$x_i$}$)$ &  What is the stance of the tweet below with respect to the probability of a merger and acquisition (M\&A) operation occurring between two companies? If the tweet is supporting the theory that the merger is happening, please label it as ‘support’. If the tweet is commenting on the merger but does not directly state that the deal is happening or refute this, please label it as ‘comment’. If the tweet is refuting that the merger is happening, please label it as ‘refute’. If the tweet is unrelated to the given merger, please label it as ‘unrelated’. Here is the tweet. ‘\textcolor{blue}{feed time... health ins. aetna to acquire humana for 37, anthem and cigna are in talks, centene corp. agreeing to buy health net \#uniteblue}’ The stance of the tweet is:\\
\midrule
PUBCLS & \textcolor{blue}{$x_i$} & \textcolor{blue}{Days after Honda announced it would add an unspecified number of vehicles with hybrid powertrains, Toyota put a number on its ambitions: 21 new hybrids in three years.} \\\\
    & $\Tcal($\textcolor{blue}{$x_i$}$)$ &  What is the publisher of the document below? You must answer even if you are not sure. If you have no clue, please make a guess. Please use exactly one word from the following 3 categories to label its entity: 'fox news', 'new york times', 'washington post'. Here is the document. ‘\textcolor{blue}{Days after Honda announced it would add an unspecified number of vehicles with hybrid powertrains, Toyota put a number on its ambitions: 21 new hybrids in three years.}’ The publisher of the mention is:\\
\midrule
SciERC & \textcolor{blue}{$x_i$} & \textcolor{blue}{machine translation pipeline [SEP] The  LOGON MT demonstrator  assembles independently valuable  general-purpose NLP components  into a  machine translation pipeline  that capitalizes on  output quality  . The demonstrator embodies an interesting combination of  hand-built, symbolic resources  and  stochastic processes.} \\\\
    & $\Tcal($\textcolor{blue}{$x_i$}$)$ &  What is the cetegory of the entity in a computer science paper abstract below? Definitions as as follows. task: Applications, problems to solve, systems to construct. method: Methods , models, systems to use, or tools, components of a system, frameworks. metric: Metrics, measures, or entities that can express quality of a system/method. material: Data, datasets, resources, Corpus, Knowledge base. others: Phrases that are a scientific terms but do not fall into any of the above classes. generic: General terms or pronouns that may refer to a entity but are not themselves informative, often used as connection words. Please use exactly one word from the following 6 categories to label its entity: 'task', 'method', 'metric', 'material', 'others', 'generic'. Here is the mention. ‘\textcolor{blue}{machine translation pipeline [SEP] The  LOGON MT demonstrator  assembles independently valuable  general-purpose NLP components  into a  machine translation pipeline  that capitalizes on  output quality  . The demonstrator embodies an interesting combination of  hand-built, symbolic resources  and  stochastic processes.}’ The entity of the mention is:\\
\bottomrule
\end{tabularx}
  \caption{Zero-Shot Prompt Templates}
\label{tab:prompt-template-zero}
\end{table*}

\section{Hyperparameters for Model Training}
\label{sec:appendix-tuning_hyperparams}

Across all the methods, the instances in the training set (labeled or pseudo-labeled) are used for training the model, and the instances (labeled or pseudo-labeled) in the validation set are used for selecting the best epoch checkpoint when training models. The instances in the testing set is held out throughout the entire training. All experiment were conducted on a GPU machine equipped with 4x NVIDIA GeForce RTX 3090. 

\subsection{Self-training Methods}
This section provides the details of the fine-tuning procedure and hyperparameters used for self-training methods (\offline, \onlinebuf, \onlinecum), as well as the baselines (\tunesource, \tunetrue). 

\paragraph{Self-train BERT-large-uncased}

Hyperparameters were determined using the validation set. We used AdamW optimizer~\cite{loshchilov2018decoupled}, along with the following hyperparameters:
\begin{itemize}
\item Learning rate: $2\times10^{-5}$
\item Weight decay: 0.01
\item Batch size: 32
\item Dropout rate: 0.1
\end{itemize}

The chosen hyperparameters were kept constant throughout the experiments to ensure fair comparisons between different model configurations.

The models were fine-tuned over a maximum of three epochs. For any $t \in [0,T]$, the fine-tuned model was selected based on the epoch checkpoint that produced the highest $F_{macro,t}$ score on the validation set.

\paragraph{Fine-tuning FLAN-T5-XXL}

Hyper parameters were determined using the validation set. We used AdamW optimizer ~\cite{loshchilov2018decoupled}, along with the following hyperparameters:
\begin{itemize}
\item Learning rate: $1\times10^{-3}$
\item Batch size: 32
\item Bottleneck dimension: 256
\item Domain classifier dimension: 1024
\item Number of epochs: 3
\item Parameter Efficient Method: LoRA 
\item LoRA rank: $16$
\item LoRA alpha ($\alpha$): $32$
\item LoRA dropout: $0.45$ 
\item Target modules: $q$, $v$
\item Bias: None
\end{itemize}

\subsection{Domain-Adversarial Training}
\label{sec:appendix-dann_hyperparams}

This subsection outlines the hyperparameters and training procedures used for the Domain-Adversarial Training (DANN) method, as described in Section~\ref{sec:method-dann}. The DANN model comprises a feature extractor and a domain classifier, where the feature extractor is based on the BERT-large architecture, along with the following hyperparameters.

\begin{itemize}[itemsep=1mm, parsep=0pt]
    \item Feature extractor: BERT-large-uncased
    \item Domain classifier and feature extractor optimizer: Stochastic Gradient Descent (SGD)
    \item Learning rate for label predictor: $1 \times 10^{-4}$
    \item Learning rate for domain classifier: $1 \times 10^{-4}$
    \item Learning rate for feature extractor: $1 \times 10^{-5}$
    \item Learning rate schedule: $lr_{i+1} = lr_{initial} \times (1 + \gamma \times i)^{-\tau}$
    \item Multiplicative factor of learning rate decay (\(\gamma\)): 0.001
    \item Exponent factor of learning rate decay (\(\tau\)): 0.75
    \item Batch size: 32
    \item Bottleneck dimension: 256
    \item Domain classifier dimension: 1024
    \item Number of epochs: 20
    \item Weight of adversarial loss (\(w_{adv}\)): 1
\end{itemize}

\subsection{Domain-Adaptive Pretraining (DAPT)}

This subsection outlines the hyperparameters and training procedures used for the DAPT method, as described in Section~\ref{sec:method-continued-pretraining}. 
DAPT leverages Flan-T5-XXL as the classification model, using AdamW~\cite{loshchilov2018decoupled} as the optimizer, along with the following training hyperparameters.

\begin{itemize}[itemsep=1mm, parsep=0pt]
    \item Text corruption rate: 15\%
    \item Learning rate: $1 \times 10^{-3}$
    \item Batch size: 32
    \item Bottleneck dimension: 256
    \item Domain classifier dimension: 1024
    \item Number of epochs: 3
    \item Parameter Efficient Method: LoRA 
    \item LoRA rank: $16$
    \item LoRA alpha ($\alpha$): $32$
    \item LoRA dropout: $0.45$ 
    \item Target modules: $q$, $v$
    \item Bias: None
    
\end{itemize}

\end{document}